\documentclass[letterpaper,10pt,conference]{ieeeconf}  

\IEEEoverridecommandlockouts                              

\usepackage{soul}
\usepackage{siunitx}
\usepackage{balance}
\usepackage{hyperref} 
\usepackage{endnotes}

\usepackage{todonotes}

\newcommand\tableheader[2]{%
  \multicolumn{1}{c}{\parbox{#1}{\centering #2}}
}

\usepackage{algorithm}
\usepackage{algorithmic}
\usepackage{empheq}
\usepackage{booktabs}
\usepackage{soul}
\usepackage{siunitx}
\usepackage{balance}
\usepackage[bottom]{footmisc}

\title{OVPC Mesh: 3D Free-space Representation for Local Ground Vehicle Navigation
}

\author{Fabio Ruetz$^{1}$, Emili Hern\'{a}ndez$^{2}$, Mark Pfeiffer$^{1}$, Helen Oleynikova$^{1}$, Mark Cox$^{2}$, Thomas Lowe$^{2}$, Paulo Borges$^{2}$  \\  
\thanks{$^{1}$Autonomous Systems Lab, ETH Zurich, Switzerland}%
\thanks{\tt\footnotesize \{ruetzf, oelena, pfmark\}@ethz.ch,}%
\thanks{$^{2}$Robotics and Autonomous Systems, Data61, CSIRO, Brisbane, Australia}%
\thanks{{\tt\footnotesize \{Name.Surname\}@csiro.au}}%
}

\begin{document}

\maketitle
\thispagestyle{empty}
\pagestyle{empty}

\begin{abstract}
This paper presents a novel approach for local 3D environment representation for autonomous unmanned ground vehicle (UGV) navigation called \emph{On Visible Point Clouds Mesh} (OVPC Mesh).
Our approach represents the surrounding of the robot as a watertight 3D mesh generated from local point cloud data in order to represent the free space surrounding the robot.
It is a conservative estimation of the free space and provides a desirable trade-off between representation precision and computational efficiency, without having to discretize the environment into a fixed grid size.
Our experiments analyze the usability of the approach for UGV navigation in rough terrain, both in simulation and in a fully integrated real-world system.
Additionally, we compare our approach to well-known state-of-the-art solutions, such as Octomap and Elevation Mapping and show that OVPC Mesh can provide reliable 3D information for trajectory planning while fulfilling real-time constraints.  

\end{abstract}

\section{Introduction}
\label{sec:introduction}
Unmanned Ground Vehicles (UGVs) have many potential applications in a broad range of industries, such as agriculture, search and rescue, and mining.
However, many of these applications require navigation in rough, unstructured, or complex terrain.
In such scenarios, 2D approaches fail to accurately represent the environment, and 2.5D maps still struggle to handle objects such as overhangs.
3D maps can accurately represent the complete environment, but usually have to be discretized to a minimum voxel size.
A coarse discretization is computationally efficient, yet does not allow for an accurate representation of small objects, non-axis-aligned surfaces, or inclines.
A fine one is computationally not tractable for real-time applications.

   \begin{figure}[tbp]
      \centering
      \includegraphics[width=\columnwidth]{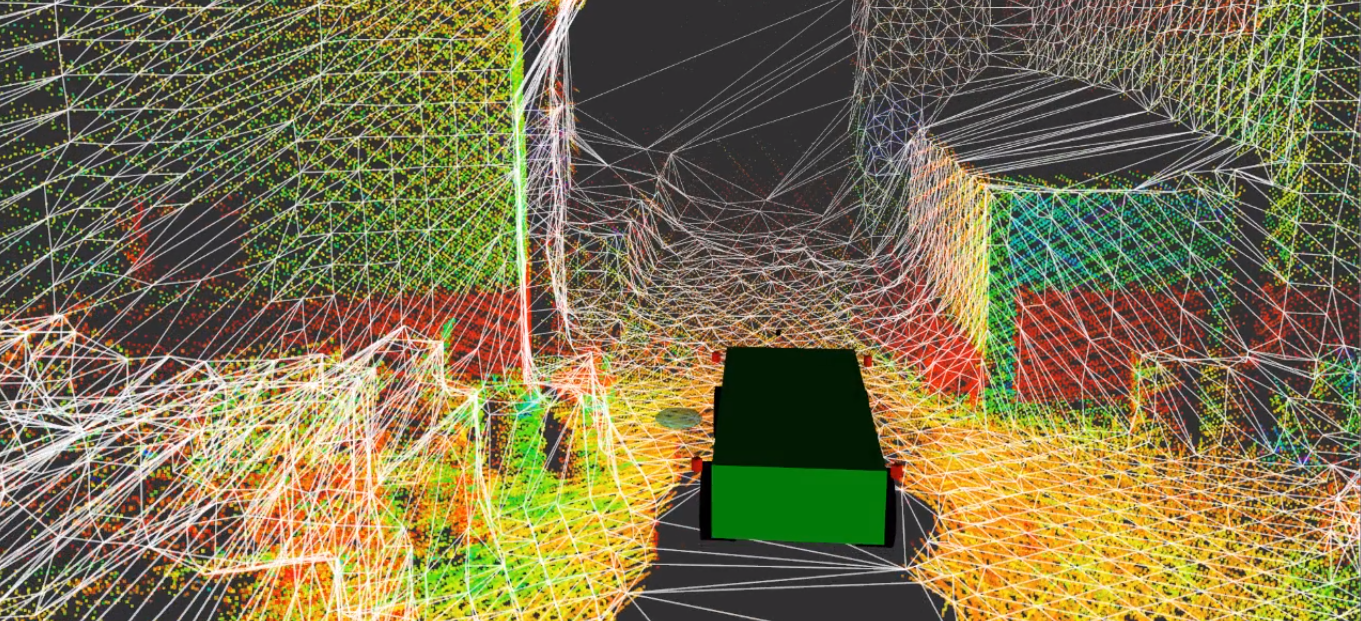}
      \caption{Raw local point cloud (colored by intensity) obtained from a 3D LiDAR and the extracted watertight mesh (white edges). 
      The mesh encapsulates the robot and gives a conservative three-dimensional free-space estimate of the local robot surrounding.
      This environment representation can also be used for rough-terrain motion planning.}
      \label{fig:teaser}
      \vspace{-4mm}
   \end{figure}
   
We present a local, robot-centric 3D map representation that explicitly models the free space around the robot to allow fast, conservative local planning.

Our approach, \emph{On Visible Point Clouds Mesh} (OVPC Mesh),  molds a watertight 3D mesh onto a point cloud simplified with the \emph{general hidden point removal operator}~\cite{Katz2015}, as shown in Figure~\ref{fig:teaser}.
Using this approach, a point cloud can be reduced to the visible points from a certain viewpoint.
The volume encapsulated by this mesh is guaranteed to be a conservative estimation of free space.
The mesh represents the information boundary, as the volume inside the mesh is free space while the outside is considered unknown.
We show how the constructed OVPC mesh can be used to estimate traversability, which can in turn be used directly in a variety of path planning techniques.

Compared to voxel-based methods, the associated computational complexity is low, which allows for a real-time deployment on robotic platforms with limited computational resources.

The capabilities of our approach are analyzed both in simulation and on a real robotic platform.
We show a comparison to existing state-of-the-art 2.5D and 3D solutions, namely Elevation Mapping~\cite{Fankhauser2016ANavigation} and Octomap~\cite{hornung2013octomap}, respectively.
Our results show successful autonomous navigation of a UGV in both structured and unstructured environments, including steep off-road inclines and overhangs. 
Our contributions can be summarized as follows:
\begin{itemize}
    \item Application of the general hidden point removal operator to generate a conservative free space representation from a raw 3D point cloud.
    \item Traversability analysis on the 3D watertight mesh for rough terrain navigation. 
    \item System evaluation on a UGV, using the traversability information for local planning purposes.
\end{itemize}

\section{Related work}
\label{sec:related_work}
A common map representation for ground vehicles is the 2D grid map~\cite{lu2014layered}. Such maps are computationally efficient and a robust solution for ground robot navigation in structured environments~\cite{marder2010office}.
2D grid maps have even been extended to navigation on rough terrain by performing terrain classification for each cell of the map~\cite{Haselich2011SplineEnvironments}.
However, since the map is still only 2D, it is very limited in what kind of environments it can represent -- and attempting to navigate in less structured environments, such as those featuring overhangs leads to either missing valid paths or executing unsafe ones.

Another common map representation for UGV operation in unstructured environments is the digital elevation map (DEM)~\cite{lacroix2002autonomous}. DEMs store a height estimate in each grid cell to capture the 2.5D geometry of the environment.
Classical DEMs suffer from two short-comings for our target applications: first, they lack the ability to represent overhangs, and second, their world-centric global formulation require accurate global pose estimates. These are often not available on real robots, where local odometry errors can accumulate to large global drifts.
Droeschel et al.~\cite{Droeschel2017ContinuousScanner} attempt to overcome the need for consistent global localization by expressing the map in a robot-centric frame. Additionally, Fankhauser et al.~\cite{Fankhauser2018}  also consider motion estimation uncertainty into their probabilistic 2.5D map.
Our approach is also robot-centric, but in contrast to these methods, is full 3D, allowing it to handle overhangs and other complex geometry.

Triebel et al.~\cite{Triebel2006} extend DEMs to global multi-level surface maps to overcome the overhangs challenge and loop-closures. They do not model overhangs explicitly, but require a minimum height distance between two surfaces to ensure a safe traversability; otherwise they are closed off. 

The easiest way to accurately represent overhangs is in a full 3D map. Many approaches exist, most of which discretize the environment into some fixed minimum voxel or block size due to the huge computational and storage requirements of such maps.
Octomap~\cite{hornung2013octomap} uses octrees to allow the system to scale to large spaces, but the octree structure requires a tree traversal for look-ups or insertions, slowing down map queries.
Voxblox~\cite{Oleynikova2017} instead uses a voxel hashing strategy, which only allocates fixed-size blocks in those regions of the map that contain data to scale to large areas without slowing down map access.
In contrast to Octomap and Voxblox, ewok~\cite{Usenko2017} uses a fixed-size robot-centric map and a ring buffer to store 3D occupancy information.
In contrast to these methods, our approach is a 3D, robot-centric map representation that does \textit{not} rely on a fixed grid-size discretization, allowing us to more accurately represent local structure of different scales.

Kr\"usi~et~al.~\cite{Krusi2017} present a 3D mapping approach specifically designed for UGV navigation on rough terrain. The map is stored as a global point cloud and the surface is reconstructed only locally for traversability analysis. Despite avoiding discretization issues, this strategy requires precise global localization and point cloud alignment to avoid global map misalignment.

Meshes are another way to represent the environment in 3D. 
Gingras et al.~\cite{Gingras2010} generate navigable spaces as triangular 3D meshes for a ground robot. These meshes represent only the surface areas the robot can traverse, excluding all non-traversable areas, which is conceptually similar to our approach. However, map updates with new sensor measurement take multiple seconds, making it unsuitable for real-time applications.
Our method demonstrates runtimes comperable to or faster than discretized methods, without trading off accuracy on small obstacles.


\section{Method}
\label{sec:methodology}

The goal of our approach is to develop a conservative 3D representation of the environment which is computationally efficient and can later be used for trajectory planning for a ground robot.
In order to reduce the overall size of the point cloud for mesh generation we use the \textit{general hidden point removal} operator (GHPR) where the \emph{visible} points from a certain viewpoint are determined.
From the sparsified point cloud, a watertight 3D triangle mesh is generated, representing the local free space.

\subsection{Mesh Generation}   
\label{subsec:method:mesh}
\begin{figure}[tbp]
      \centering
      \includegraphics[width = 0.9\columnwidth]{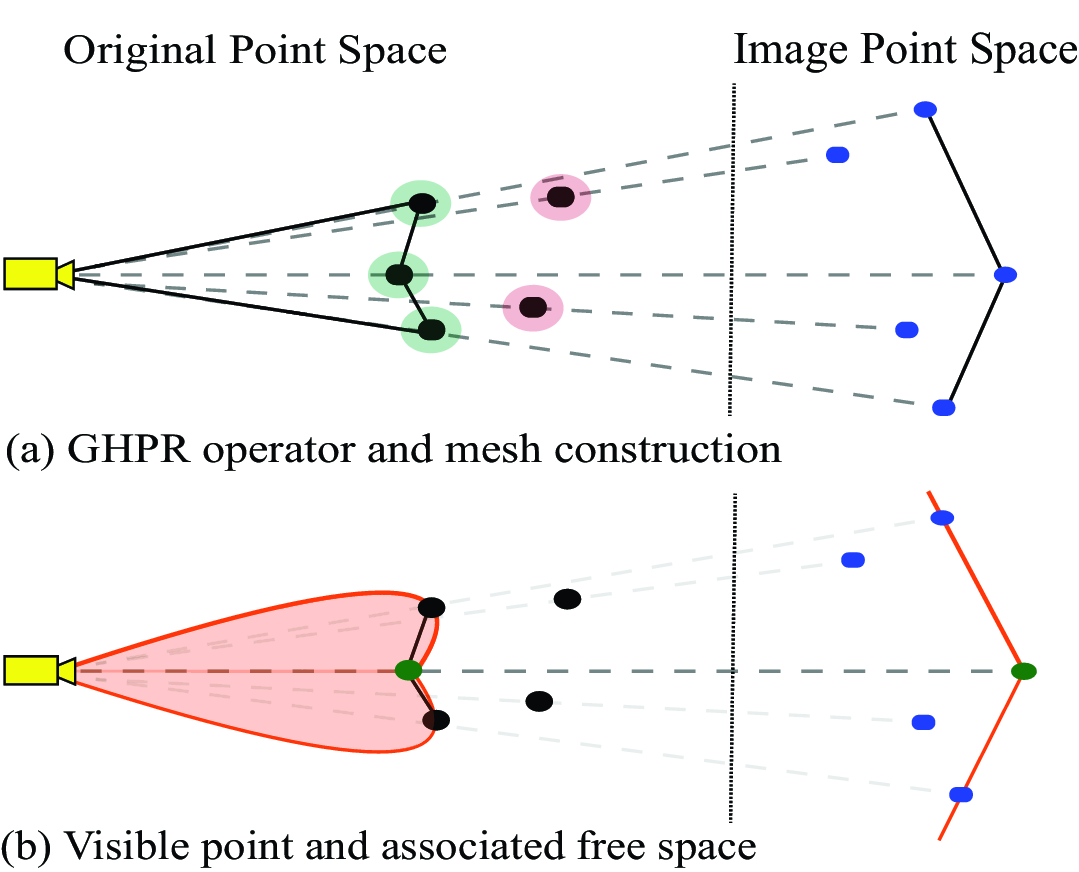}
      \caption{
      Mesh generation based on the GHPR algorithm: (a) Transformation to generate the image points (blue) from the original point set (black). Convex hull over image points generates connectivity between the points in the image space and can be applied to the original point set(black full lines). Points in the original space are visible (green transparent) if their images lie on the convex hull, else they are considered non-visible (red transparent). (b) Extending the edges from a visible point (green) to its neighbors to infinity in the are mapped to curves in the original point space. These curves bound an area or a volume in 2D or 3D (transparent orange), and are guaranteed to be free of other observations.
      }
      \label{fig:ghpr}
      \vspace{-4mm}
   \end{figure}
In order make the meshing operation real-time capable, we use GHPR, which is commonly used in computer graphics to find visible points on objects from a given viewpoint, without explicitly reconstructing surfaces.
As shown in Figure~\ref{fig:ghpr}, the original point set (OPS), which is obtained from a local point cloud, is transformed to an image point set (IPS) using a radial transformation (see Figure~\ref{fig:ghpr}).
Making use of the transformation, the visible point classification is reduced to the problem of forming a convex hull over the (transformed) IPS.
A point is considered visible if its image lies on the convex hull in the image space. 

As Figure~\ref{fig:ghpr} shows, the connectivity information of the convex hull in the image space can be transferred to the original space in order to construct a triangle mesh on the visible points.
The resulting mesh is closed and watertight, and it only depends on the visible points (from the current viewpoint) of the OPS.

\begin{figure}[tbp]
      \centering
      \includegraphics[width=\columnwidth]{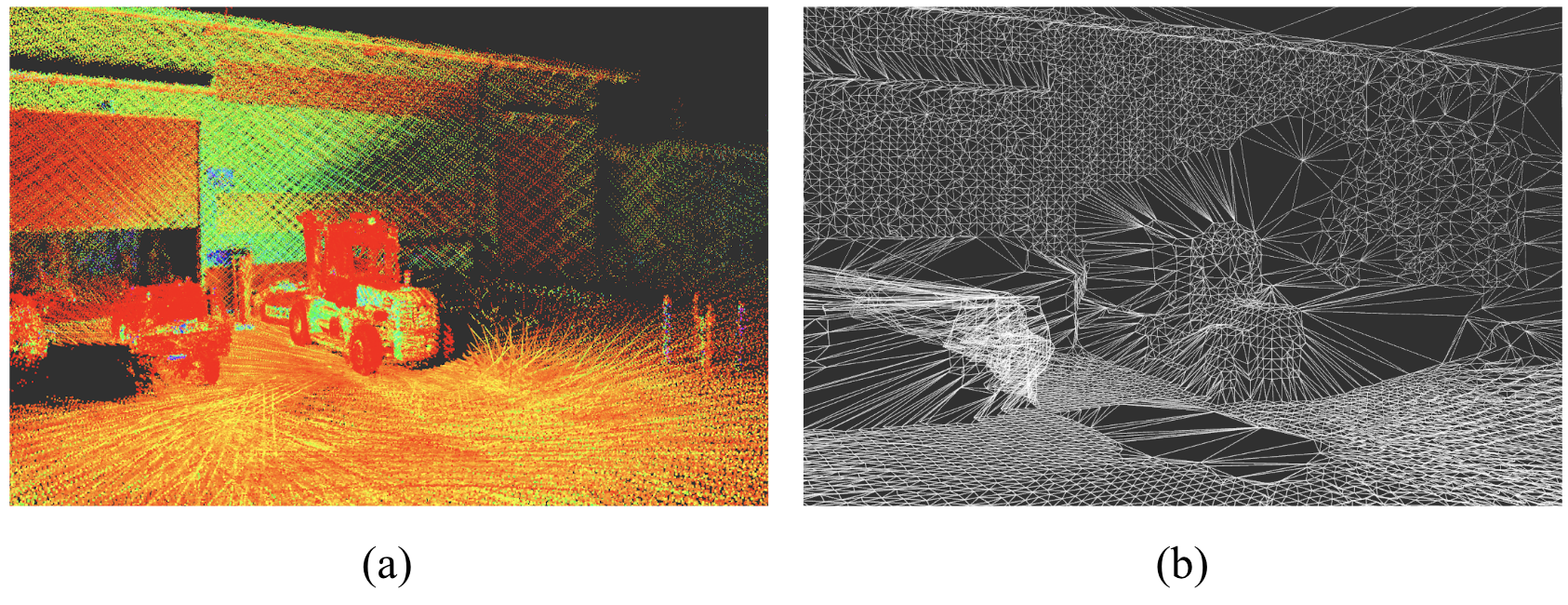}
    \caption{Example of an OVPC Mesh from point cloud obtained with our UGV: (a) intensity colored point cloud; (b) corresponding watertight triangle mesh generated from the point cloud. }
    \vspace{-2mm}
      \label{fig:pcl_to_mesh}
   \end{figure}

By construction, this method solves the visibility of point classification dependant on free space associated with each point in the OPS; free from any other points. The shape is controlled by the kernel and limited by the point's neighbours (Figure~\ref{fig:ghpr}(b)). The kernel parameter $\gamma$ defines the minimum required space for a point to be visible. This results in all visible points lying on the boundary region free of other observations. We use the exponential kernel and given the shape of the curves forming the boundary, and we assume that given two visible points their connecting edge will lie within free space.
Under the assumptions of perfect sensor readings and all edges lying in free space, the generated mesh is a conservative bound of the free space surrounding the robot. The interested reader should refer to~\cite{Katz2015} for more detailed explanation. 

\begin{figure}[tbp]
      \centering
      \includegraphics[width=\columnwidth]{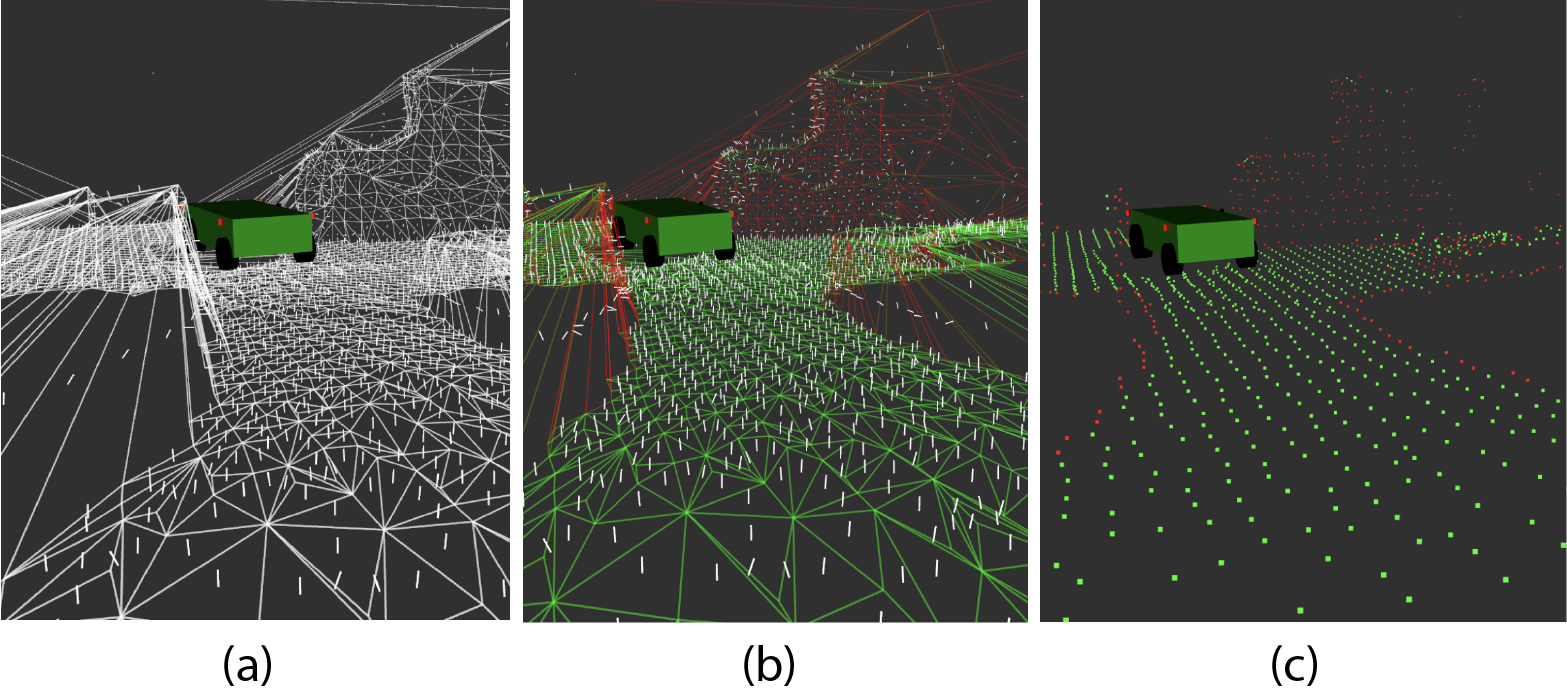}
      \caption{Assessing the traversability on the mesh faces and generating a map for navigation. Figure (a) shows the generated water-tight mesh and the normal estimation of each face as short white line segments, (b) shows each face classified in traversable (green) and non-traversable (red) using geometric constraints described in Section~\ref{subsec:method:planning} and (c) shows the extracted point cloud representation where each point is classified based on its traversability. 
      Note that the outsides of the traversable area are strictly non-traversable which allows for conservative navigation.}
      \label{fig:trav}
      \vspace{-4mm}
   \end{figure}

\subsection{Map Representation for Path Planning} 
\label{subsec:method:planning}

Once the mesh is generated, we perform a traversability analysis on each polygon to constrain the local free space according to our vehicle specifications. For this analysis, we use the surface angle $\alpha$: this is the angle between the surface normal and the gravity-aligned world frame. A mesh polygon is \emph{traversable} if: the surface angle is smaller than the threshold $\alpha_{\max}$ and lies within the interval [-$\pi/2$ , $\pi/2$]; and the maximum height difference among its vertices is smaller than $\delta_{h,\max}$. Otherwise the polygon is set as \emph{non-traversable}.

Inspired by~\cite{Krusi2017}, a point cloud is formed from the vertices of the mesh. Each point is assigned a normal vector, representing the estimated surface, and a binary traversability score.  
The normal is calculated as mean of all normals from the adjacent faces.  A point is only considered traversable if all the neighboring faces are traversable, see Figure~\ref{fig:trav}(c). This point cloud is used as a map representation for local planning.

\section{Results}
\label{sec:results}

In this section we describe our specific hardware and software implementation followed by the results in simulation, the ones obtained with a fully integrated UGV and a comparison of our approach against two state-of-the-art methods. A video of the proposed method can be found under the following link~\footnote{ \href{https://youtu.be/8b0w56bg0WM}{https://youtu.be/8b0w56bg0WM} }. 

\subsection{Hardware and Software Setup}

A John Deere Gator automated by CSIRO~\cite{egger2018posemap} was used as the test platform. 
The vehicle is equipped with a Velodyne PUCK VLP-16 LiDAR and a Microstrain GX3 IMU. 
The LiDAR is mounted at a height of \SI{1.88}{\meter} above the robot frame at a \SI{45}{\degree} angle, as illustrated in the bottom right of Figure~\ref{fig:qcat}. 
The LiDAR assembly is additionally mounted on a spinning base, which rotates at approximately \SI{0.5}{\hertz}~\cite{egger2018posemap}, and LiDAR measurements are streamed in at \SI{20}{\hertz}.

Our mesh approach requires the LiDAR sensor readings and the local pose transforms~\cite{egger2018posemap} for our pipeline.
No global localization is needed during the mesh construction and the local path computation steps.

  \begin{figure}[tbhp]
      \centering
      \includegraphics[width=\columnwidth]{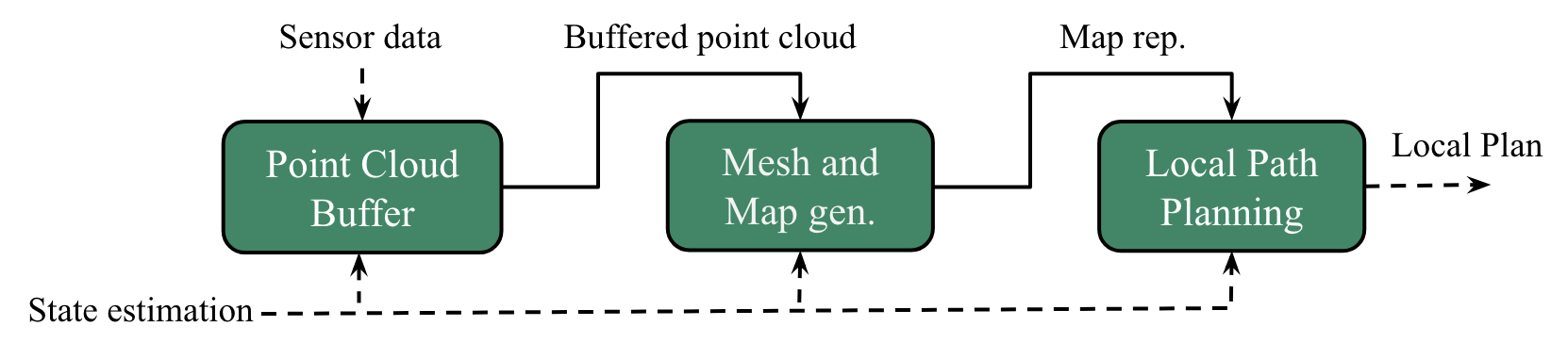}
    \caption{Navigation pipeline overview: the sensor data is accumulated in the point cloud buffer, filtered and sub-sampled; then, the mesh and map generation module computes the watertight triangle mesh using the GHPR operator, which is then used for traversability analysis. Finally, the local planner generates a trajectory from the map, which is passed on to a trajectory follower. 
    }
      \label{fig:overview_OVPC_pipeline}
      \vspace{-4mm}
   \end{figure}

The system was implemented in C++ using ROS~\cite{quigley2009ros}, running on an external Dell Precision 7520 (Intel i7-4910MQ CPU@\SI{2.90}{\giga\hertz} and 16$GB$ of RAM) connected via Ethernet. 
An overview of our processing pipeline is depicted in Figure~\ref{fig:overview_OVPC_pipeline} and consists of the following three steps:

\subsubsection{Point cloud buffer} 
\label{subsec:point-clooud-buffer}
In the first step a point cloud buffer aligns, filters and sub-samples the data to generate a common point cloud from a fixed number (20-30) of previous scans. 
In our tests, the resulting scans contain around 6,000 to 14,000 points depending on the environment.

\subsubsection{Mesh and map generation} 
\label{subsec:mesh-gen}
A watertight triangle mesh is then generated around the robot and the traversability of the faces is evaluated. The mesh vertices are used to compute the visible point cloud used for path planning, where each point contains a normal and traversability information. This step is described in detail in Section~\ref{subsec:method:mesh} and~\ref{subsec:method:planning}.

\subsubsection{Local path planning}
\label{subsec:local-plan}
We used the RRTconnect~\cite{Kuffner2000} sampling based path planner from OMPL~\cite{sucan2012the-open-motion-planning-library} in combination with the Reeds-Shepp~\cite{Soueres1996} state space to generate a local plan given a local goal. 
The local goal is obtained from a given global plan and has to lie within the mesh at a minimum distance ahead of the robot. 
During planning, each sampled state in $SE(2)$ is projected onto the map using the closest visible point in Euclidean space. The state is aligned to the local surface using the normal of the visible point, while preserving the heading. 
Collisions are detected by checking all visible points within a bounding box according to the platform dimensions, which is aligned with the sampled pose.
Re-planning is enforced if (i) the current path is in collision after a map update, (ii) the current goal is too close to the current pose of the robot or (iii) a time threshold has been exceeded (\SI{1}{\second}). 
The maximum computation time for planning is set to \SI{0.1}{\second}.
If successful, the local path is passed on to a trajectory follower~\cite{Giesbrecht2005}.

\subsection{Evaluation in simulation}
\label{sec:simulations}
Merging multiple point cloud scans together can lead to misalignments.
Thus, we explore how our method performs when merging noisy, misaligned scans and how well the underlying geometry can be described by using our method instead of the raw point clouds. 
An accurate estimation of the geometry is important, as the traversability estimation depends on the normals of the mesh faces.
We benchmark our approach against an other normal estimation technique~\cite{RusuDoctoralDissertation} based on raw point clouds. 
We use the angular error to evaluate the accuracy and variance between the estimated normal and ground truth.
This was evaluated in a $\SI{20}{\meter} \times \SI{20}{\meter}$ scene with a flat ground plane and an inclined plane with a defined slope $\alpha$.
We vary $\alpha$ between \SI{0}{\degree} and \SI{35}{\degree} with \SI{0.1}{\degree} increments to determine if the inclination has any effect on our normal estimation.

To simulate noisy bundled LiDAR measurements, we bundle 5 simulated point clouds together, each containing points spaced \SI{0.2}{\meter} apart and $\pm\SI{0.05}{\meter}$ uniform noise in each axis.
A uniform error in orientation of $\pm \SI{0.5}{\degree}$ and translation of \SI{0.1}{\meter} was added to the ground and slope, representing the pose uncertainty. 
Each scene is evaluated for a given angle $\alpha$. 
The robot was positioned in the centre of the ground plane with the viewpoint at \SI{1.88 }{\meter} above ground (LiDAR mount).

As a comparison we use the standard normal estimation method~\cite{RusuDoctoralDissertation} available in PCL~\cite{Rusu_ICRA2011_PCL} to estimate the normal of a point. A radius of $r = \SI{0.3}{\meter}$ was used to perform the nearest neighbour search. 

We use the angular error as the performance metric, i.e. the angular distance between the estimated and the ground truth normals.
  \begin{figure}[thpb]
     \centering
     \includegraphics[width=\columnwidth,trim={0 8mm 0 0}]{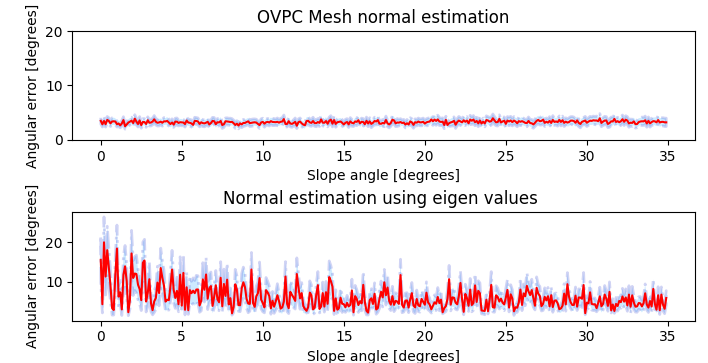}
     \caption{Graphs of angular error of different normal estimation methods on noisy, misaligned point clouds. Mean angular error (red) and $\pm \sigma$ (blue). \textbf{Top:} The angular error is consistent, mean of \SI{3.25}{\degree} over all slope angles. The variance is also lower comparing to the method using eigenvalues. \textbf{Bottom:} The method using eigenvalues shows a higher angular error on average and higher variance. 
     }
     \label{fig:eval_normals}
     \vspace{-6mm}
  \end{figure}
  
The top plot in Figure.~\ref{fig:eval_normals} shows our method mean angular error of \SI{3.25}{\degree} over the full range of slopes. The variance is bound and consistent for all tested slope angles. 
In practice this error can be observed by our traversability estimation when classifying traversable points as non-traversable. This can be mainly observed on the ground plane, where small triangles are oriented in such a way that they close the mesh to make it watertight which potentially results in steep inclines.

The eigenvalue method~ \cite{RusuDoctoralDissertation} shows significantly worse results for the angle estimates given the same noisy point cloud data.
It shows a higher angular error and variance. 
Our method uses the local surroundings of a mesh surface to generate the normal estimations.
We adjusted the radius for nearest neighbor search of the eigenvalue method to match the size of this area to obtain fair results.
Increasing the number of nearest neighbours or the coverage radius will make the estimation smoother and reduce the variance in this scenario, but will lead to biased angle estimation in the presence of more obstacles and complex environments. 
Also, as shown in \cite{Krusi2017}, it requires additional computations on the point cloud to capture discontinuities and surface roughness. OVPC Mesh captures those situations implicitly due to coverage area used for the surface estimation.
In essence this experiment shows that our proposed method is capable of accurately estimating the local geometry using noisy point cloud scans.

\subsection{Experimental results}
\label{sub:real_world_exp}
\begin{figure*}[thpb]
      \centering
      \includegraphics[width=\textwidth]{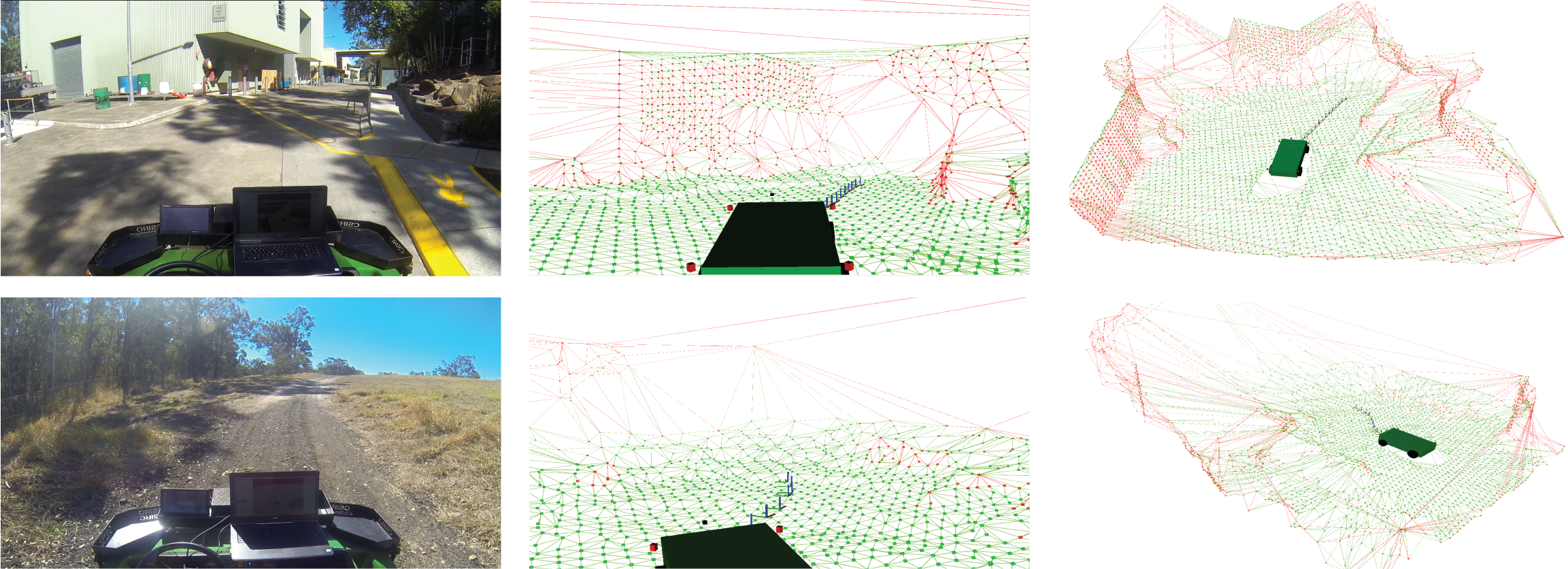}
      \caption{The left column depicts the view from an on board camera during the autonomous navigation in structured (\textbf{top}) and unstructured  (\textbf{bottom}) environments. The middle column shows the same scene but from the view of the robot inside of the mesh. The faces of the mesh are classified as traversable (green) and non-traversable (red). The local plan can be seen as a series of axes extending from the front of the robot. The right column shows the same scene from outside of the mesh and visualizes the size of the area considered for local navigation.
    }
      \label{fig:long_run}
      \vspace{-4mm}
   \end{figure*}

In addition to the simulation experiments, which validate that our approach can be used for accurate geometry representation and traversability estimation, we also conduct a real-world evaluation of the fully integrated system.
We conducted three fully autonomous long runs with an total distance of more than \SI{1.5}{\kilo\meter}. 
These runs capture a variety of different environments at the Queensland Centre of Advanced Technology (QCAT) in Brisbane, ranging from structured industrial sites to unstructured slopes on steep inclines. 
The different scenarios are also shown in Figure \ref{fig:long_run}.
Parameters used for the experiments are chosen as follows: $\gamma = -0.03$, $\delta_{h,\max} = \SI{0.25}{\meter}$,  $\alpha_{\max} = \SI{30}{\degree}$, the viewpoint coincides with the LiDAR mount position and the velocity was set at $1 \frac{\text{m}}{\text{s}}$.
In general, the map representation performed well in structured and semi-structured environments, which were encountered on all 3 runs. Thin obstacles, such as posts,  small trees, or walls were accurately detected and avoided while the global plan was followed. The robot successfully navigated through these areas without any human intervention.

The end of yellow and red run consisted of a steep slope in a unstructured area. Driving up this slope introduced severe vibrations into the system. In general, this led to an increase in noise but the traversability analysis was still robust enough for navigation. In several instances, strong bumps and the resulting errors in the state estimation were observed to increase the noise such that the traversability estimation failed. This forced the robot to stop and re-plan in order to find a new trajectory. 
This resulted in a jerky movement of the robot on some parts of the slopes.  
Note that these issues appear only with the fully integrated system when navigating in rough terrain and were not introduced by our OVPC Mesh method.

\begin{figure}[thpb]
      \centering
      \includegraphics[width=0.7\columnwidth]{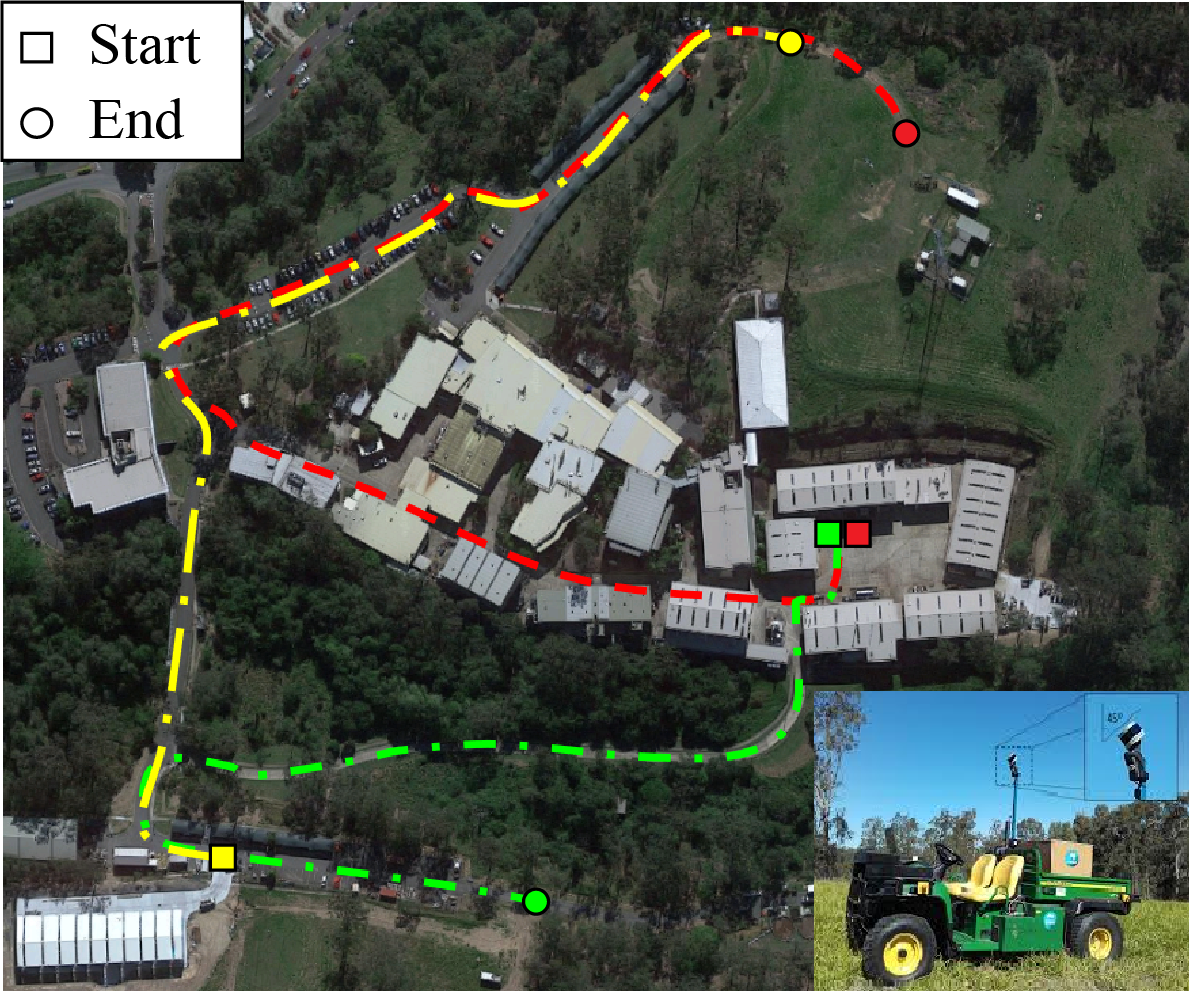}
      \caption{Three different fully autonomous runs performed at QCAT. The overall covered distance is more than \SI{1.5}{\kilo\meter} and includes both structured and off-road environments. \textbf{Bottom right:} The John Deere Gator test platform. 
      The LiDAR sensor is mounted at a \SI{45}{\degree} angle and on a rotating base, allowing the robot to observe more of its environment even with a limited vertical field of view.}
      \label{fig:qcat}
      \vspace{-6mm}
   \end{figure}

\begin{table}[]
\centering
\caption{Statistics of the runs shown in in Figure~\ref{fig:qcat}.}
\begin{tabular}{cccc}
   \toprule
    \tableheader{1.5cm}{\textbf{Run}} &  \tableheader{1.5cm}{\textbf{Time} {[}s{]}} &  \tableheader{2.0cm}{\textbf{Distance} {[}m{]}} &
    \tableheader{1.5cm}{\textbf{Color}} \\
    \midrule[1.5pt] 
    Run 1 & 542     & $\sim$605  & Red  \\
    Run 2 & 503       & $\sim$425  & Yellow   \\
    Run 3 & 520       & $\sim$480 & Green  \\
    \bottomrule
    \end{tabular}
  \vspace{-8mm}
\end{table}

\subsection{Comparison to other methods}   
We compare our method to two map representations commonly used for robot path planning: Octomap~\cite{hornung2013octomap}, a full 3D representation, and Elevation Mappping~\cite{Fankhauser2016ANavigation}, a 2.5D height map. 
The purpose is to evaluate our method in terms of map representation quality and computational performance in challenging environments for UGV navigation.

\subsubsection{Runtime benchmarking}
We compare the runtime of our method to Octomap and Elevation Mapping, using a single bundled point cloud to generate a map suitable for planning. These point clouds are the result of our point cloud buffer on real-world data sets and were randomly chosen from the runs shown in Figure~\ref{fig:qcat}.

\begin{figure}[thb]
      \centering
      \includegraphics[width=\columnwidth,trim={0 8mm 0 8mm}]{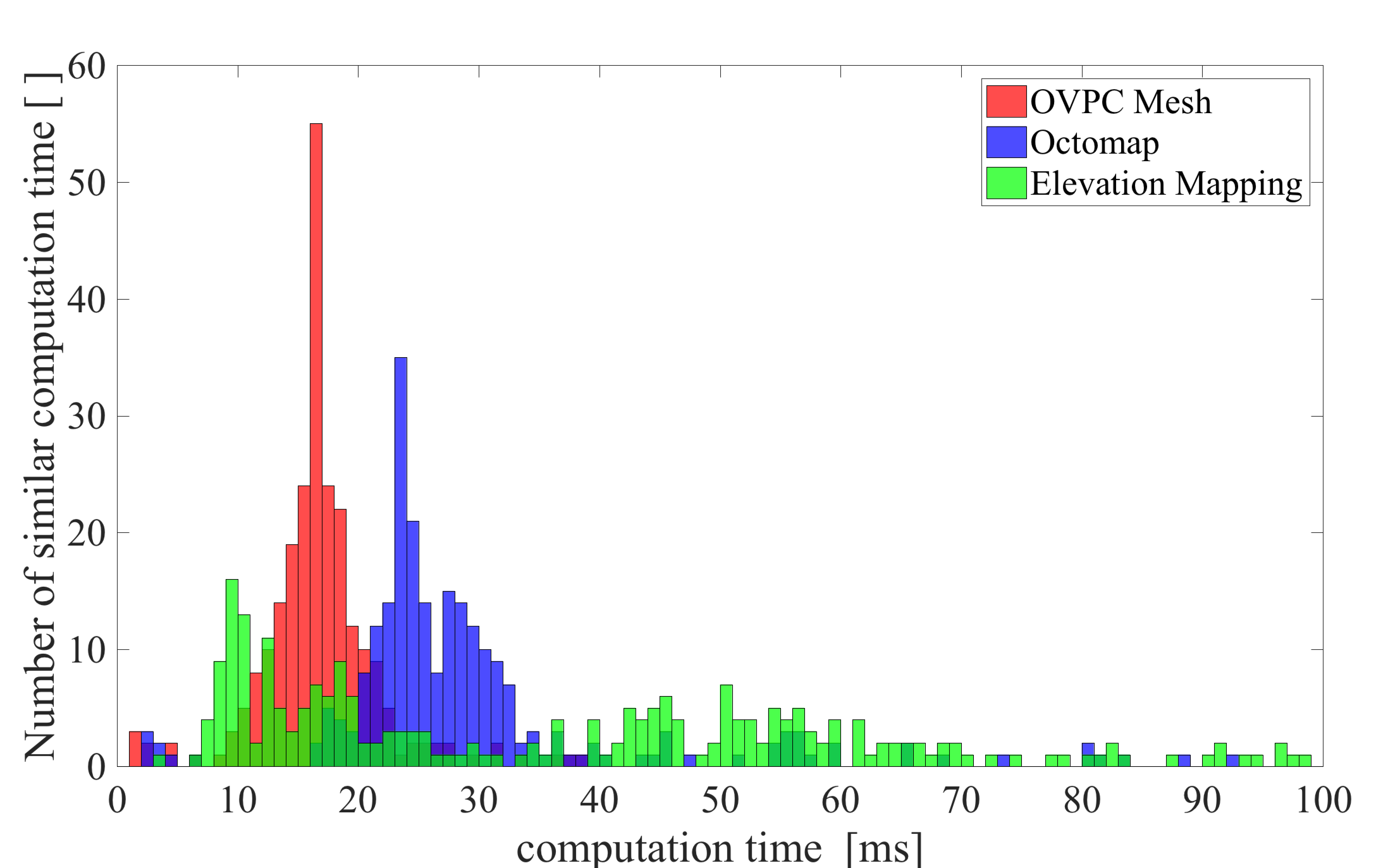}
     \caption{Single point cloud map generation computation time histograms: OVPC Mesh (red) requires a mean time of $16.78 \pm 5.04$ms. Octomap (blue) requires $30.91 \pm 18.66$ms to integrate the same data. Elevation Mapping (green) generates the fused map in $36.68 \pm 25.10$ms.}
      \label{fig:bench_ovpc_octo}
      \vspace{-6mm}
\end{figure}

Figure~\ref{fig:bench_ovpc_octo} shows the runtime distributions of OVPC Mesh, Octomap and Elevation Mapping. Our method is not only faster on average ($\mu=\SI{16.78}{\milli\second}$ (our method) vs. $\SI{30.91}{\milli\second}$ (Octomap) vs. $\SI{36.68}{\milli\second}$ (Elevation Mapping)) but also has a lower spread and therefore more predictable runtimes ($\sigma=\SI{5.04}{\milli\second}$ vs. $\SI{18.66}{\milli\second}$ vs. $\SI{25.10}{\milli\second}$), which is especially desirable for fully integrated real-world systems.

\subsubsection{Handling of overhangs}
Overhanging obstacles express a challenge for UGV navigation pipelines.
In rough terrain this requires a simultaneous traversability analysis of the ground and an analysis of the overhangs.
Since the LiDAR sensor is mounted at an exposed position at the very top of the vehicle, we need to pay particular attention to overhangs for autonomous navigation of our system.

\begin{figure}[thpb]
      \centering
      \includegraphics[width=\columnwidth,trim={0 8mm 0 0}]{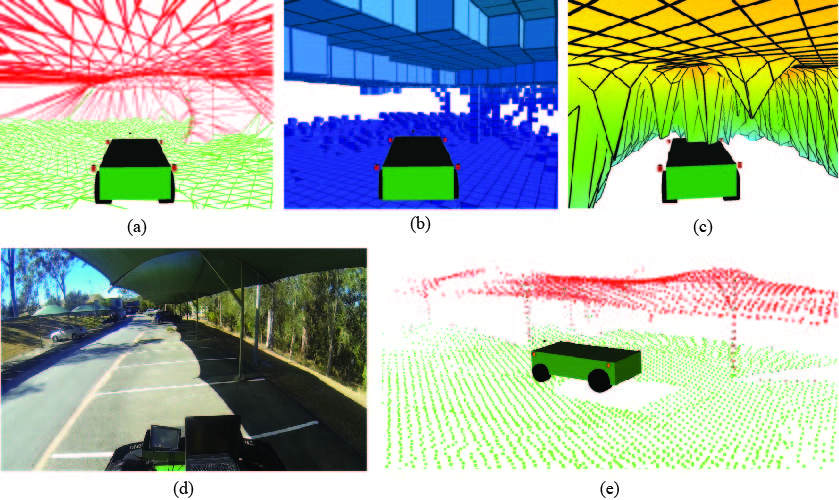}
      \caption{A comparison of how different mapping techniques handle overhangs, shown from the UGV on-board camera in (d).
      (a) shows our method, with the mesh colored by non-traversable (red) and traversable (green), shown from a different perspective in (e), where thin obstacles, poles and thin trees, are accurately classified. The mesh extends from under the obstacle, allowing for local navigation. In (b), Octomap captures the sun cover and ground accurately. In (c), Elevation Mapping struggles with overhang, roof and floor outside of the overhang are reconstructed but not under the overhang.}
      \label{fig:overhang}
      \vspace{-6mm}
   \end{figure}

Figure~\ref{fig:overhang}~(d) shows an example of an overhanging structure at the QCAT car park, which has sunroofs above the parking spaces. Figure~\ref{fig:overhang}~(a) depicts our 3D world representation with traversable faces colorized in green and non-traversable in red. 
The roof and ground are clearly separated w.r.t. traversability and the mesh extends from under the roof, allowing for local navigation. This is further illustrated in a wide angle shot (Figure~\ref{fig:overhang}~(e)). Additionally, the figure also shows the robustness of our approach in terms of capturing thin objects such as poles and small trees, and its traversability classification in a challenging environment. 

Figure~\ref{fig:overhang}~(b) shows the 3D voxel representation of the overhang scenario using Octomap. It captures the overhang accurately and also extends from under the overhang.  
During the \SI{86}{\second} run the Octomap's ray integration slowed down to an update rate lower than \SI{1}{\hertz} after \SI{50} and \SI{60}{\second} using voxel sizes between \SI{0.3} and \SI{0.5}{\meter} respectively. 
No better results were obtained by further parameter tuning.

In addition to Octomap, we tested Elevation Mapping in the overhang test scenario.
The result using this 2.5D approach is depicted in Figure~\ref{fig:overhang}~(c). Despite being capable of modeling the area not covered by the roof sufficiently well, the area underneath the roof is not modeled at all. We assume that this is caused by the point exclusion heuristic, which operates based on the point height and distance to the sensor. Adjusting the tuning parameters could not mitigate this effect in this particular environment since the sensor height is \SI{1.88}{\meter}, which is closer to the roof than to the ground in this application. 

\section{Discussion}

OVPC Mesh has shown to be suitable for local UGV navigation in structured and unstructured environments. 
However, there are still several open challenges which are listed below. 
Although more robust than raw point clouds, the mesh generation process is sensitive to outliers in the LiDAR data, which can make the mesh unusable for planning. Applying a voxel filter, minimum 2 measurements per voxel to generate a point, has shown to be sufficient to remove the most extreme outliers. 

Furthermore, the kernel parameter $\gamma$, which determines the free space required for a point to be visible, has a strong effect on the mesh generation (see Section~\ref{subsec:mesh-gen}). 
Empirical evaluation showed that values $\gamma \in [-0.01, -0.03]$, in combination with the exponential kernel~\cite{Katz2015}, generate the most suitable results for UGV navigation in the environments we target in this work.
When choosing the kernel values to close to zero  nearly all points become visibible and are connected by edges, as little space is required per point to be visible. Larger values will result in a sparser mesh.
The values chosen should be in accordance with the spacing of the points in the point cloud and is application and sensor dependent.

Using the convex hull algorithm to compute a triangular mesh ensures that it is watertight. As a drawback, this process also generates triangles that do not represent the environment but connect points at the boundaries, where no LiDAR data is provided. 
However, such blind spots (like negative obstacles) are a common issue linked to local sensing and it is beyond the scope of this paper.

Lastly, the mesh is viewpoint dependent and any change on it will require the generation of a new mesh. 
Consequently, the path generated by the local path planner is only valid within the encapsulated volume. 
Hence, fast mesh re-generation and path re-planning are required, but we believe that the mean $\SI{16}{\milli\second}$ runtime of our method is suitable for most applications.

\section{Conclusions}
\label{sec:conclusions}
In this paper, we presented a novel method to build a local 3D free space representation, and showed its applicability in UGV rough terrain navigation.

We generated a local map consisting of a watertight, closed mesh which is a conservative representation of the visible free space. We additionally computed the traversability of each mesh face and generate a map representation suitable for local path planning according to our vehicle constraints. 

We validated the accuracy of our traversability estimation in simulation, even in the presence of significant sensor noise.
We demonstrated our approach through autonomous navigation on a fully integrated real system, navigating through structured, semi-structured, and unstructured environments, also including challenges like low hanging ceilings and rough off-road terrains with steep inclines.
Lastly, we compared our approach to two state-of-the-art solutions both qualitatively and quantitatively.
The results show that our approach provides precise environment representations while being computationally efficient compared to existing methods.

\addtolength{\textheight}{-3cm}

\bibliographystyle{IEEEtran}
\balance
\bibliography{bibliography}

\end{document}